\pdfoutput=1

\documentclass[11pt]{article}

\usepackage[final]{acl}

\usepackage{times}
\usepackage{latexsym}

\usepackage[T1]{fontenc}

\usepackage[utf8]{inputenc}

\usepackage{microtype}

\usepackage{inconsolata}

\usepackage{graphicx}
\usepackage{makecell} 
\usepackage{multirow, array}

\usepackage{booktabs}

\usepackage{threeparttable}

\usepackage{comment}

\newcommand{\corrauthor}{\textsuperscript{\dag}}

\title{OR-Toolformer: Modeling and Solving Operations Research Problems with Tool Augmented Large Language Models}

 \author{Jianzhang Zhang, Jialong Zhou \and Chuang Liu\corrauthor \\
         Alibaba Business School, Hangzhou Normal University\\
         \texttt{\{zjzhang,liuchuang\}@hznu.edu.cn}, \texttt{jialongzhouzj@gmail.com}}

\begin{document}
\maketitle

\begingroup
\renewcommand\thefootnote{\dag}
\footnotetext{Corresponding author.}
\endgroup

\begin{abstract}
Large language models (LLMs) demonstrate strong mathematical reasoning, but reliance on closed-source APIs for OR tasks raises privacy concerns, and training open-source models from scratch incurs high compute costs. We introduce OR-Toolformer, which fine-tunes Llama-3.1-8B-Instruct with a semi-automatic data synthesis pipeline that generates diverse OR problem-answer pairs and augments the model with external solvers to produce API calls. On three of four standard benchmarks, OR-Toolformer achieves up to 80.1\% execution accuracy, exceeding size-matched baselines by over 4.3\%. In zero-shot evaluation on two unseen OR problem types, it attains 54\% average accuracy, a 21 percentage-point improvement over the strongest baseline. These findings validate the efficacy of tool-augmented fine-tuning LLMs for accurate and generalizable OR problem modeling and solving.
\end{abstract}

\section{Introduction}

Operations Research (OR) offers rigorous methods to formalize and solve complex decision problems in various sectors. OR workflows involve (1) translating natural-language descriptions into mathematical optimization models and (2) obtaining solutions via general-purpose solvers~\cite{Petropoulos2024}, yet this pipeline remains dependent on domain expertise, limiting scalability.

Large language models (LLMs) have demonstrated strong text comprehension and multi-step mathematical reasoning on complex benchmarks~\cite{RomeraParedes2024,Xia2025}, indicating their potential to automate both formulation and solution of OR tasks. However, reliance on closed source LLM APIs raises data privacy concerns~\cite{Das2025}, as sensitive problem descriptions and data often constitute commercial confidential information and must be transmitted to proprietary platforms beyond the user’s control. Moreover, training open-source models from scratch incurs prohibitive computational costs~\cite{Xia2024}.

Fine-tuning pre-trained LLMs for domain-specific tasks offers a resource-efficient alternative, but vanilla LLMs struggle with precise arithmetic~\cite{McLeish2024}. Tool-learning techniques enable LLMs to invoke external tools, such as calculators or specialized APIs, thereby combining generative flexibility with solver accuracy~\cite{schick2023toolformer,Shi2025}. We introduce OR-Toolformer~\footnote{publicly available after finishing peer reviewing}, which fine-tunes Llama-3.1-8B-Instruct to extract structured solver parameters from natural-language OR problem descriptions and generate corresponding API calls, fully automating the modeling and solution phases.







\begin{figure}[h]
	\centering
	\includegraphics[width=\linewidth]{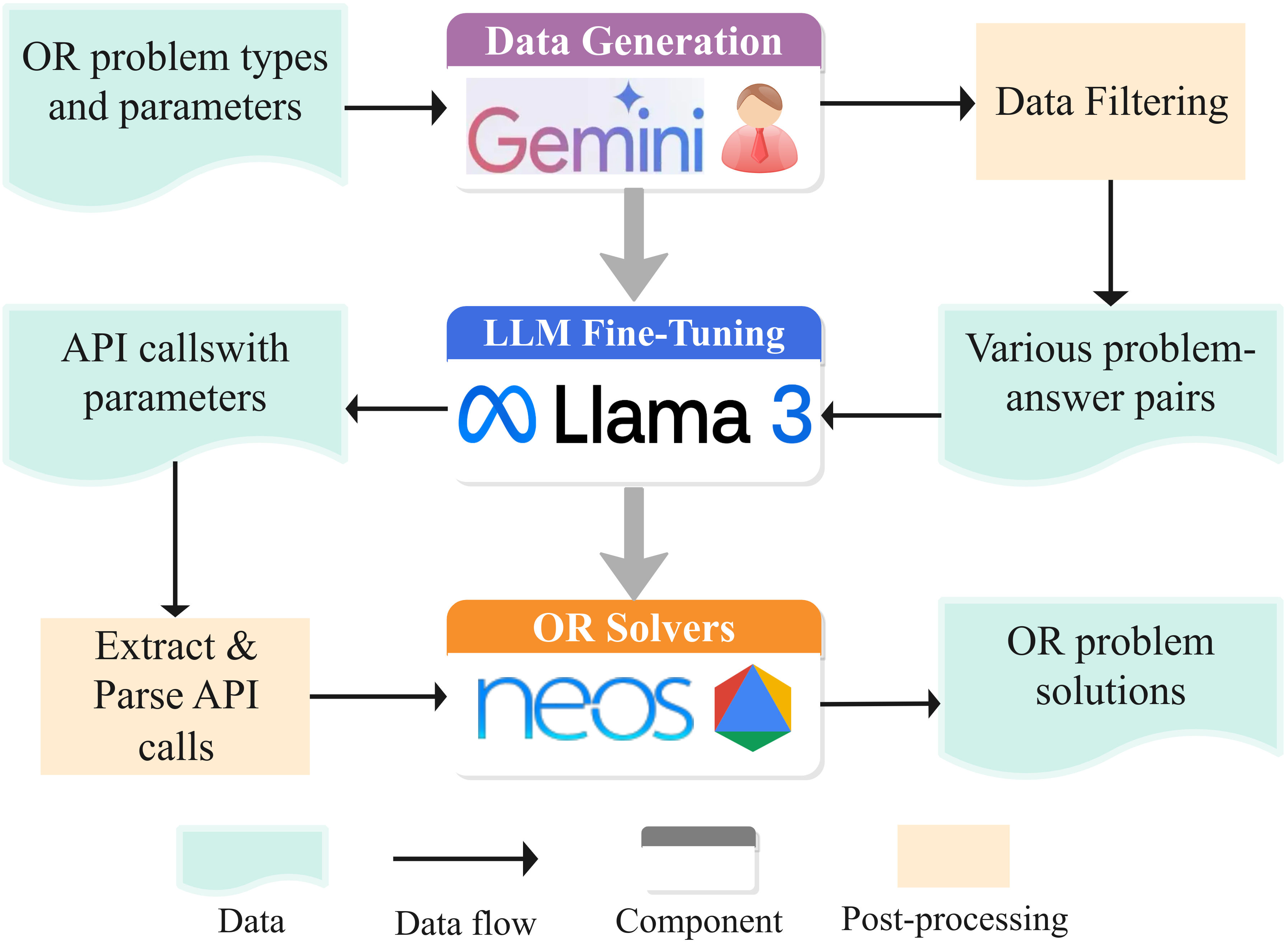}
	\caption{Overview of OR-Toolformer.}
	\label{fig:overview}
\end{figure}

\section{The Methodology of OR-Toolformer}

OR-Toolformer automates OR tasks through three integrated components (Figure~\ref{fig:overview}): 
\begin{itemize}
	\item \textbf{Problem–Answer Data Generation}, a semi-automated pipeline that synthesizes diverse OR problem-answer pairs across problem types, industry contexts, and representation formats to ensure domain and expression diversity;
	
	\item \textbf{LLM Fine-Tuning}, which adapts pre-trained LLMs to parse natural-language descriptions and extract structured solver parameters;
	
	\item \textbf{Problem Solving with OR Solvers}, where the fine-tuned model issues API calls to external optimization solvers, uniting language comprehension with computational precision.
\end{itemize}

%
%
%


\subsection{Problem-Answer Data Generation}

\begin{figure}[h]
	\centering
	\includegraphics[width=\linewidth]{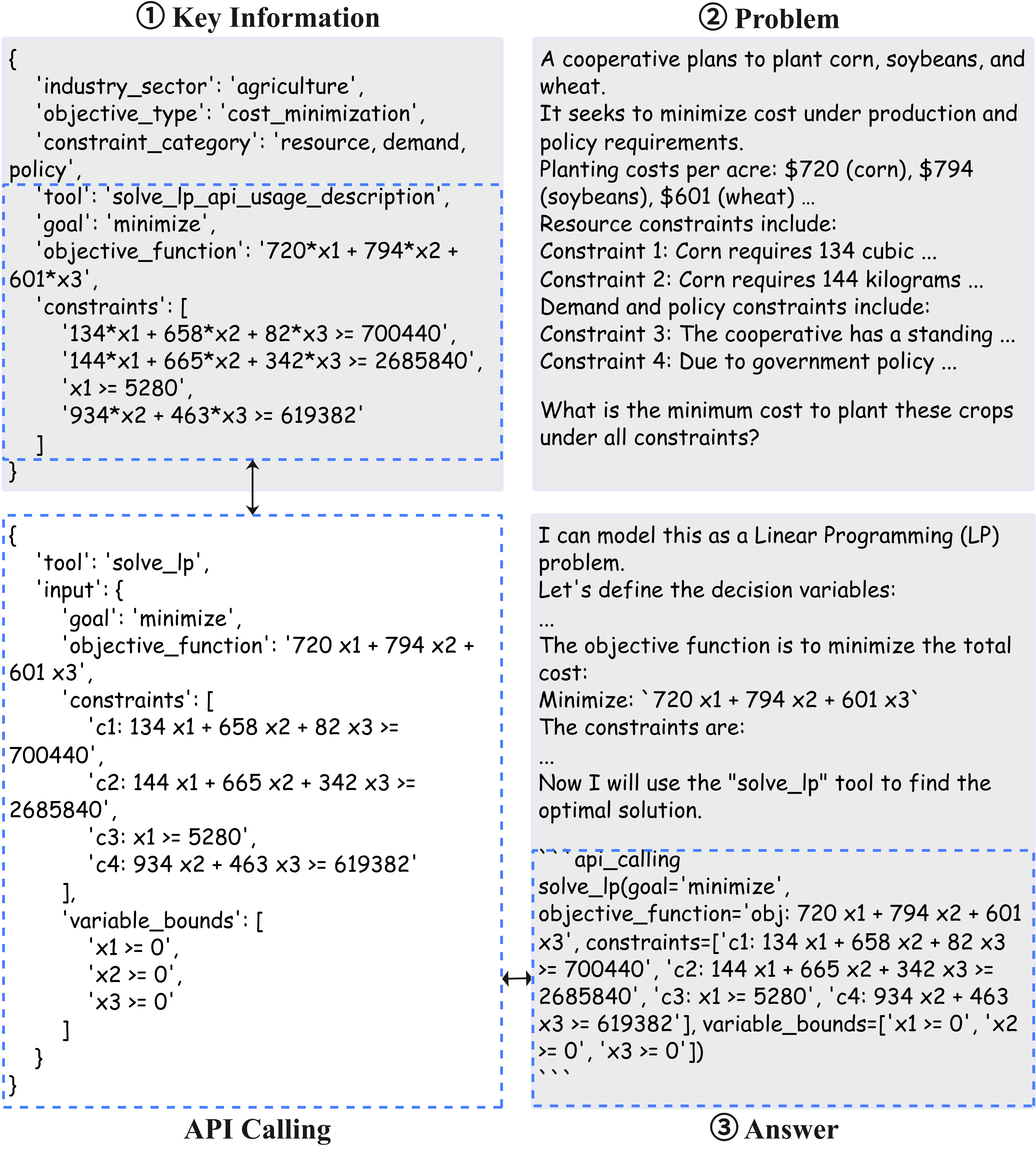}
	\caption{Snippet of the generation process of an LP problem-answer pair.}
	\label{fig:data-generation}
\end{figure}

High-quality instruction tuning for robust generalization requires OR problem-answer pairs that capture both domain-specific variation and diverse linguistic expressions~\cite{Albalak2024}. Given the scarcity of datasets that include detailed modeling steps and solver API calls~\cite{Huang2025a,Mostajabdaveh2025}, we introduce a three-stage, semi-automated pipeline for large-scale synthesis of OR problem-answer pairs. Figure~\ref{fig:data-generation} presents an example linear programming (LP) problem-answer instance, highlighting the input key information (left-top) and the generated problem-answer pair (right) along with the corresponding API call (left-bottom).


\textbf{Stage~1: Parameter sampling.} We randomly sample OR problem parameters from realistic ranges (e.g., positive unit consumption rates). These values are converted into structured API inputs and validated by the solvers. To ensure \emph{domain diversity}, we vary application contexts (e.g., agriculture, logistics, finance) and objective types (profit maximization, cost minimization). For \emph{expression diversity}, the parameter set of OR problem is rendered in free-form text, matrix notation, and tabular lists.


\textbf{Stage~2: Prompt-based statement and answer synthesis.} We embed the key information (the sampled parameters and context as illustrated in top-left of Figure~\ref{fig:data-generation}) into a problem generation prompt template that instructs Gemini 2.0 Flash to generate coherent OR problem statements. We then augment the same key information with API usage descriptions in answer generation prompt template that instructs Gemini 2.0 Flash to produce both the chain of thoughts and the corresponding API call (bottom‐right of Figure~\ref{fig:data-generation}). These two prompts are shown in Appendix \ref{sec:question-prompt} and \ref{sec:answer-prompt} respectively.

\textbf{Stage~3: Quality filtering and formatting.} To mitigate hallucinations~\cite{Huang2025}, we execute the generated API call (dotted box in the right-bottom of Figure~\ref{fig:data-generation}) and compare its result against that of the sampled parameters based API call (left bottom of Figure~\ref{fig:data-generation}).  Only problem-answer pairs with matching results are retained. Finally, we cast validated instances into a dialogue format aligning with instruction‐tuning best practices~\cite{ouyang2022training,Qin2024,patil2024gorilla}. System messages list one correct tool and three distractors, user messages present the problem, and assistant messages deliver the chain of thoughts plus API calls.

\subsection{LLM Fine-Tuning}

We fine-tune \textit{Llama-3.1-8B-Instruct}~\cite{grattafiori2024llama} on our synthesized dataset via instruction tuning. Let \(\mathcal{D} = \{(Q_i, A_i)\}_{i=1}^N\) denote the set of \(N\) OR problem–answer pairs, where each prompt \(Q_i\) comprises a system message and a user message, and \(A_i\) is the corresponding assistant message. The model's prediction for \(Q_i\) is \(\hat{A}_i = \mathrm{LLM}_\theta(Q_i)\). We optimize the parameters \(\theta\) by minimizing the negative log-likelihood (cross-entropy) loss:
\begin{equation}
	\mathcal{L}(\theta) = -\sum_{i=1}^N \log P_\theta(A_i \mid Q_i)
\end{equation}
where \(P_\theta(A_i \mid Q_i)\) is the probability assigned by the LLM to the reference output \(A_i\).

%

\subsection{Problem Solving with OR Solvers}


After generating an OR problem solution, we extract the embedded API call strings and parse them into structured invocations as depicted by the connected dotted boxes in the bottom of Figure~\ref{fig:data-generation}. We execute these on two external OR services, NEOS Server\footnote{https://neos-server.org/neos/} and Google Operations Research API\footnote{https://developers.google.com/optimization/service}, to compute numerical solutions for each instance\footnote{Google OR API is used to solve MF, MCF, and AP problems, as NEOS does not offer solvers for these problems.}. 



\section{Experiments}

%
%

\subsection{Experimental Setup}

\textbf{Data generation.} We synthesize two datasets: one for instruction fine‐tuning and another to evaluate zero‐shot generalization on unseen OR problem types. Table~\ref{tbl:data} details the number of instances per problem category. The fine-tuning dataset includes the same types of problems as those found in the four benchmarks including NL4OPT~\cite{ramamonjison2023nl4opt}, MAMO-EasyLP, MAMO-ComplexLP~\cite{huang2024mamo}, and IndustryOR~\cite{Huang2025a}.

\begin{table}[htbp]
	\centering
	\small
	
	\begin{tabular}{cccccc}
		\toprule
		\multicolumn{6}{c}{\textbf{Training Dataset}} \\
		\midrule
		LP & IP & MILP & TSP & MF & \textbf{Total} \\
		3502 & 3501 & 3493 & 3516 & 3496 & 17508 \\
		\midrule
		\multicolumn{6}{c}{\textbf{Test Dataset}} \\
		\midrule
		TSP & MF & AP & MCF & & \\
		50 & 50 & 50 & 25 & & 175 \\
		\bottomrule
	\end{tabular}
	\vspace{1mm}
	
	\parbox{0.86\columnwidth}{\footnotesize \textbf{Abbreviations:} LP = Linear Programming; IP = Integer Programming; MILP = Mixed-Integer Linear Programming; TSP = Traveling Salesman Problem; MF = Maximum Flow; AP = Assignment Problem; MCF = Minimum-Cost Flow.}
	
	\caption{Summary statistics of training and test datasets.}
	\label{tbl:data}
\end{table}



\textbf{Training.} We fine-tune \textit{Llama-3.1-8B-Instruct} on the full training set with a batch size of 64 and a learning rate of $2\times10^{-4}$. Using the Unsloth~\cite{unsloth} framework on a single GPU (10 GB VRAM), we perform parameter-efficient fine-tuning via LoRA, 8-bit AdamW, and 4-bit quantization, updating only 0.52\% of parameters.



\textbf{Evaluation.} We measure execution accuracy following Huang et al.~\cite{Huang2025a}, deeming a prediction correct if the solver's returned optimum matches any ground‐truth value. We benchmark OR-Toolformer against general‐purpose LLMs (ChatGPT, Gemini, DeepSeek-R1) and size‐matched baselines: general LLMs (DeepSeek-7B, Mistral-7B, Qwen-2.5-7B) and math-focused LLMs (JiuZhang-3.0).

\subsection{Results Analysis}

\begin{table*}[!htbp]
	\centering
	\small
	
	\begin{tabular}{clccccc}
		\toprule
		\multicolumn{2}{c}{\textbf{Method}} & \textbf{NL4OPT} & \makecell{\textbf{MAMO-}\\ \textbf{EasyLP}} & \makecell{\textbf{MAMO-}\\ \textbf{ComplexLP}} & \textbf{IndustryOR} \\
		\midrule
		
		\multirow{4}{*}{\textbf{General LLMs}} & GPT-3.5 & 42.4\% & 61.8\% & 20.9\% & 19.0\% \\
		& GPT-4 & 47.3\% & 66.5\% & 14.6\% & 28.0\% \\
		& Gemini-2.0 Flash & 79.6\% & 77.3\% & 26.1\% & 23.0\% \\
		& DeepSeek-R1-685B & 66.1\% & 73.6\% & 48.3\% & 27.0\% \\
		\midrule
		
		\multirow{4}{*}{\shortstack{\textbf{General LLMs} \\ \textbf{in similar scale}}} & DeepSeek-LLM-7B-Chat & 5.7\% & 2.3\% & 0.5\% & 1.0\% \\
		& Llama-3.1-8B-Instruct & 6.9\% & 8.3\% & 7.6\% & 3.0\% \\
		& Mistral-7B-Instruct-v0.3 & 0.0\% & 0.0\% & 0.0\% & 3.0\% \\
		& Qwen-2.5-7B-Instruct & \underline{44.1\%} & \underline{43.6\%} & 9.5\% & \textbf{18.0\%} \\
		\midrule
		
		\multirow{5}{*}{\shortstack{\textbf{Math LLMs} \\ \textbf{in similar scale}}} & DeepSeek-Math-7B-Instruct & 20.0\% & 30.7\% & 6.6\% & 10.0\% \\
		& DeepSeek-Math-7B-RL & 23.7\% & 27.5\% & 10.4\% & 10.0\% \\
		& Qwen-2.5-Math-7B & 40.8\% & 41.1\% & \underline{10.9\%} & 9.0\% \\
		& JiuZhang-3.0-7B & 13.9\% & 4.6\% & 3.3\% & 4.0\% \\
		& JiuZhang-3.0-8B & 23.7\% & 4.3\% & 4.3\% & 2.0\% \\
		\midrule
		
		\textbf{Ours} & OR-Toolformer-8B & \textbf{59.6\%} & \textbf{80.1\%} & \textbf{14.7\%} & \underline{14.0\%} \\
		\bottomrule
	\end{tabular}\\
	\vspace{1mm}
	
	\parbox{0.86\textwidth}{\footnotesize \textbf{Note.} The best results are in bold, and the second-best are underlined. Results in the first section are not included in the ranking.}
	\caption{Performance of OR-Toolformer and three types of baselines on four benchmarks.}
	\label{tbl:performance}
\end{table*}

\textbf{Results on benchmarks.} Table~\ref{tbl:performance} summarizes the execution accuracy of OR-Toolformer and baseline models on four standard benchmarks. All models achieve substantially higher accuracy on simpler tasks (NL4OPT, MAMO-EasyLP) than on more complex ones (MAMO-ComplexLP, IndustryOR). Consistent with scaling laws ~\cite{Kaplan2020}, larger general-purpose LLMs outperform their smaller counterparts on three of the four benchmarks. Accordingly, we focus our analysis on size-matched general-purpose and math-specific LLMs. Among 7-8 B models, OR-Toolformer delivers the highest accuracy across all benchmarks except IndustryOR, where it places second. In particular, OR-Toolformer attains 80.1\% on MAMO-EasyLP and approximately 14\% on both MAMO-ComplexLP and IndustryOR, substantially outperforming other size-matched baselines, all of which fall below 18\%. Although Qwen-2.5-7B-Instruct ranks second, math-specific LLMs generally outperform other size-matched models, underscoring the value of domain-specific fine-tuning~\cite{Zhang2024}.

\begin{table}[h]
	\centering
	\small
	
	\begin{tabular}{>{\raggedright}m{2cm}*{4}{>{\centering\arraybackslash}p{0.9cm}}}
		\toprule
		
		Method & TSP & MF & AP & MCF \\
		\midrule
		Qwen-2.5-7B-Instruct & 0.0\% & 16.0\% & 62.0\% & 4.0\% \\
		\midrule
		Ours & \textbf{100.0\%} & \textbf{98.0\%} & \textbf{68.0\%} & \textbf{40.0\%} \\
		\bottomrule
	\end{tabular}
	
	\caption{Performance of OR-Toolformer and Qwen-2.5-7B-Instruct on test dataset.}
	\label{tbl:zero-shot}
\end{table}



\textbf{Results on the test dataset.} Table~\ref{tbl:zero-shot} compares OR-Toolformer and Qwen-2.5-7B-Instruct on two OR problem types (AP and MCF) not included in the benchmark suites. On two familiar problem types (TSP and MF), which were generated identically to our training data, OR-Toolformer achieves 100\% and 98\% execution accuracy, respectively, confirming the consistency of our synthesis pipeline. Crucially, on two entirely unseen problem types (AP and MCF), OR-Toolformer attains 68\% and 40\% accuracy versus 62\% and 4\% for Qwen-2.5-7B-Instruct, representing an average improvement of 21 percentage points. These results demonstrate OR-Toolformer's strong zero-shot generalization to novel OR tasks.



\textbf{Output token efficiency.} We evaluate the average output length of each model to assess token efficiency. OR-Toolformer generates concise responses, averaging 449 tokens, compared to 500 tokens for Qwen-2.5-7B-Instruct and 1,422 tokens for Qwen-2.5-Math-7B, the latter of which typically includes extensive mathematical derivations and embedded code. As illustrated in the bottom-right of Figure~\ref{fig:data-generation}, OR-Toolformer produces succinct natural-language outputs that satisfy both optimization and API-invocation requirements, thereby substantially reducing token consumption.

\section{Related Work}

Tool learning enables LLMs to extend generative capacity by invoking external APIs. STE has models imagine, execute, and refine tool-usage sequences via simulated trial-and-error~\cite{Wang2024}. Cooperative multi-agent methods decompose tool use into grounding, execution, and review stages~\cite{Shi2024}, and budget-constrained planning generates cost-optimal call sequences under resource limits~\cite{Zheng2024}. Self-instruction pipelines synthesize diverse API-call examples from documentation~\cite{Yang2023}, further scaled by Shi et al.\cite{Shi2025}. Large-scale benchmarks such as StableToolBench\cite{Guo2024} and RoTBench~\cite{Ye2024} standardize evaluation, and ToolSword exposes safety vulnerabilities across tool-learning stages~\cite{Ye2024a}. Unlike prior work focused on calculator-based tools~\cite{schick2023toolformer}, our method emphasizes solver learning for OR, leveraging self-instruction generated training data~\cite{Yang2023}.

LLMs have been applied to automate OR task formulation and solution. The NL4OPT competition provides a widely used benchmark~\cite{ramamonjison2023nl4opt}, and Mostajabdaveh et al.\cite{Mostajabdaveh2025} evaluate open-source LLMs on complex OR problems. Chain-of-Experts and Optimus combine prompt engineering and multi-agent pipelines using GPT-4 for OR formulation~\cite{Xiao2023,AhmadiTeshnizi2024}. LLMs have also been used to help interpret optimization results and identify infeasible optimization problems~\cite{li2023large,Chen2024}. To mitigate privacy and computational costs, ORLM fine-tunes open-source models end-to-end for solver-code generation~\cite{Huang2025a}. In contrast, we employ parameter-efficient fine-tuning to yield concise natural language formulations and structured API calls.


\section{Conclusion}



We present OR-Toolformer, a fine-tuned Llama-3.1-8B-Instruct model augmented with external OR solvers. It achieves 80.1\% execution accuracy on three standard benchmarks, outperforming size-matched LLMs, and 54\% average zero-shot accuracy on two novel problem types (a 21 pp improvement). These results confirm the efficacy of tool-augmented LLM fine-tuning for both accuracy and generalization in OR tasks. Future work will explore integrating agents via a model-context protocol.

\clearpage
\clearpage

\section*{Limitations}

Our study has several limitations. First, OR-Toolformer's accuracy on complex or industry-scale OR tasks (e.g., MAMO-ComplexLP, IndustryOR) remains substantially lower than on simpler academic benchmarks, which may impede real-world deployment. Second, due to computational constraints, we fine-tuned and evaluated only a single open-source LLM; a broader comparison across additional models is left to future work. Third, our synthetic data pipeline relies on heuristic prompt templates and limited domain context; incorporating stronger LLMs and richer industrial scenarios could enhance data realism and diversity. Finally, we have not yet conducted user-centered evaluations to measure the framework's usability and utility in practical optimization workflows.

\bibliography{custom}

\begin{thebibliography}{32}
\providecommand{\natexlab}[1]{#1}

\bibitem[{AhmadiTeshnizi et~al.(2024)AhmadiTeshnizi, Gao, and
  Udell}]{AhmadiTeshnizi2024}
Ali AhmadiTeshnizi, Wenzhi Gao, and Madeleine Udell. 2024.
\newblock \href {https://openreview.net/forum?id=YT1dtdLvSN} {Optimus: scalable
  optimization modeling with (mi) lp solvers and large language models}.
\newblock In \emph{Proceedings of the 41st International Conference on Machine
  Learning}, pages 577--596.

\bibitem[{Albalak et~al.(2024)Albalak, Elazar, Xie, Longpre, Lambert, Wang,
  Muennighoff, Hou, Pan, Jeong, Raffel, Chang, Hashimoto, and
  Wang}]{Albalak2024}
Alon Albalak, Yanai Elazar, Sang~Michael Xie, Shayne Longpre, Nathan Lambert,
  Xinyi Wang, Niklas Muennighoff, Bairu Hou, Liangming Pan, Haewon Jeong, Colin
  Raffel, Shiyu Chang, Tatsunori Hashimoto, and William~Yang Wang. 2024.
\newblock \href {https://openreview.net/forum?id=XfHWcNTSHp} {A survey on data
  selection for language models}.
\newblock \emph{Transactions on Machine Learning Research}.
\newblock Survey Certification.

\bibitem[{Chen et~al.(2024)Chen, Constante-Flores, and Li}]{Chen2024}
Hao Chen, Gonzalo~E Constante-Flores, and Can Li. 2024.
\newblock \href {https://doi.org/10.1080/03155986.2024.2385189} {Diagnosing
  infeasible optimization problems using large language models}.
\newblock \emph{INFOR: Information Systems and Operational Research},
  62(4):573--587.

\bibitem[{Das et~al.(2025)Das, Amini, and Wu}]{Das2025}
Badhan~Chandra Das, M~Hadi Amini, and Yanzhao Wu. 2025.
\newblock \href {https://doi.org/10.1145/3712001} {Security and privacy
  challenges of large language models: A survey}.
\newblock \emph{ACM Computing Surveys}, 57(6):1--39.

\bibitem[{Grattafiori et~al.(2024)Grattafiori, Dubey, Jauhri, Pandey, Kadian,
  Al-Dahle, Letman, Mathur, Schelten, Vaughan et~al.}]{grattafiori2024llama}
Aaron Grattafiori, Abhimanyu Dubey, Abhinav Jauhri, Abhinav Pandey, Abhishek
  Kadian, Ahmad Al-Dahle, Aiesha Letman, Akhil Mathur, Alan Schelten, Alex
  Vaughan, and 1 others. 2024.
\newblock \href {https://doi.org/10.48550/arXiv.2407.21783} {The llama 3 herd
  of models}.
\newblock \emph{arXiv preprint arXiv:2407.21783}.

\bibitem[{Guo et~al.(2024)Guo, Cheng, Wang, Liang, Qin, Li, Liu, Sun, and
  Liu}]{Guo2024}
Zhicheng Guo, Sijie Cheng, Hao Wang, Shihao Liang, Yujia Qin, Peng Li, Zhiyuan
  Liu, Maosong Sun, and Yang Liu. 2024.
\newblock \href {https://doi.org/10.18653/v1/2024.findings-acl.664}
  {{S}table{T}ool{B}ench: Towards stable large-scale benchmarking on tool
  learning of large language models}.
\newblock In \emph{Findings of the Association for Computational Linguistics:
  ACL 2024}, pages 11143--11156, Bangkok, Thailand. Association for
  Computational Linguistics.

\bibitem[{Han and Han(2023)}]{unsloth}
Daniel Han and Michael Han. 2023.
\newblock \href {http://github.com/unslothai/unsloth} {Unsloth}.

\bibitem[{Huang et~al.(2025{\natexlab{a}})Huang, Tang, Hu, Jiang, Zheng, Ge,
  Wang, and Wang}]{Huang2025a}
Chenyu Huang, Zhengyang Tang, Shixi Hu, Ruoqing Jiang, Xin Zheng, Dongdong Ge,
  Benyou Wang, and Zizhuo Wang. 2025{\natexlab{a}}.
\newblock \href {https://doi.org/10.1287/opre.2024.1233} {Orlm: A customizable
  framework in training large models for automated optimization modeling}.
\newblock \emph{Operations Research}.

\bibitem[{Huang et~al.(2025{\natexlab{b}})Huang, Yu, Ma, Zhong, Feng, Wang,
  Chen, Peng, Feng, Qin et~al.}]{Huang2025}
Lei Huang, Weijiang Yu, Weitao Ma, Weihong Zhong, Zhangyin Feng, Haotian Wang,
  Qianglong Chen, Weihua Peng, Xiaocheng Feng, Bing Qin, and 1 others.
  2025{\natexlab{b}}.
\newblock \href {https://doi.org/10.1145/3703155} {A survey on hallucination in
  large language models: Principles, taxonomy, challenges, and open questions}.
\newblock \emph{ACM Transactions on Information Systems}, 43(2):1--55.

\bibitem[{Huang et~al.(2024)Huang, Shen, Hu, Gao, and Wang}]{huang2024mamo}
Xuhan Huang, Qingning Shen, Yan Hu, Anningzhe Gao, and Benyou Wang. 2024.
\newblock \href {https://doi.org/10.48550/arXiv.2405.13144} {Mamo: a
  mathematical modeling benchmark with solvers}.
\newblock \emph{arXiv preprint arXiv:2405.13144}.

\bibitem[{Kaplan et~al.(2020)Kaplan, McCandlish, Henighan, Brown, Chess, Child,
  Gray, Radford, Wu, and Amodei}]{Kaplan2020}
Jared Kaplan, Sam McCandlish, Tom Henighan, Tom~B Brown, Benjamin Chess, Rewon
  Child, Scott Gray, Alec Radford, Jeffrey Wu, and Dario Amodei. 2020.
\newblock \href {https://arxiv.org/abs/2001.08361} {Scaling laws for neural
  language models}.
\newblock \emph{arXiv preprint arXiv:2001.08361}.

\bibitem[{Li et~al.(2023)Li, Mellou, Zhang, Pathuri, and Menache}]{li2023large}
Beibin Li, Konstantina Mellou, Bo~Zhang, Jeevan Pathuri, and Ishai Menache.
  2023.
\newblock \href {https://doi.org/10.48550/arXiv.2307.03875} {Large language
  models for supply chain optimization}.
\newblock \emph{arXiv preprint arXiv:2307.03875}.

\bibitem[{McLeish et~al.(2024)McLeish, Bansal, Stein, Jain, Kirchenbauer,
  Bartoldson, Kailkhura, Bhatele, Geiping, Schwarzschild et~al.}]{McLeish2024}
Sean McLeish, Arpit Bansal, Alex Stein, Neel Jain, John Kirchenbauer, Brian
  Bartoldson, Bhavya Kailkhura, Abhinav Bhatele, Jonas Geiping, Avi
  Schwarzschild, and 1 others. 2024.
\newblock \href
  {http://papers.nips.cc/paper\_files/paper/2024/hash/c35986bc1ee29b31c1011481b77fe540-Abstract-Conference.html}
  {Transformers can do arithmetic with the right embeddings}.
\newblock In \emph{Advances in Neural Information Processing Systems}, pages
  108012--108041.

\bibitem[{Mostajabdaveh et~al.(2025)Mostajabdaveh, Yu, Dash, Ramamonjison,
  Byusa, Carenini, Zhou, and Zhang}]{Mostajabdaveh2025}
Mahdi Mostajabdaveh, Timothy Tin~Long Yu, Samarendra Chandan~Bindu Dash, Rindra
  Ramamonjison, Jabo~Serge Byusa, Giuseppe Carenini, Zirui Zhou, and Yong
  Zhang. 2025.
\newblock \href {https://doi.org/10.1609/aaai.v39i23.34673} {Evaluating llm
  reasoning in the operations research domain with orqa}.
\newblock In \emph{Proceedings of the AAAI Conference on Artificial
  Intelligence}, pages 24902--24910.

\bibitem[{Ouyang et~al.(2022)Ouyang, Wu, Jiang, Almeida, Wainwright, Mishkin,
  Zhang, Agarwal, Slama, Ray et~al.}]{ouyang2022training}
Long Ouyang, Jeffrey Wu, Xu~Jiang, Diogo Almeida, Carroll Wainwright, Pamela
  Mishkin, Chong Zhang, Sandhini Agarwal, Katarina Slama, Alex Ray, and 1
  others. 2022.
\newblock \href
  {http://papers.nips.cc/paper\_files/paper/2022/hash/b1efde53be364a73914f58805a001731-Abstract-Conference.html}
  {Training language models to follow instructions with human feedback}.
\newblock In \emph{Advances in Neural Information Processing Systems}, pages
  27730--27744.

\bibitem[{Patil et~al.(2024)Patil, Zhang, Wang, and
  Gonzalez}]{patil2024gorilla}
Shishir~G Patil, Tianjun Zhang, Xin Wang, and Joseph~E Gonzalez. 2024.
\newblock \href
  {http://papers.nips.cc/paper\_files/paper/2024/hash/e4c61f578ff07830f5c37378dd3ecb0d-Abstract-Conference.html}
  {Gorilla: Large language model connected with massive apis}.
\newblock In \emph{Advances in Neural Information Processing Systems}, pages
  126544--126565.

\bibitem[{Petropoulos et~al.(2024)Petropoulos, Laporte, Aktas, Alumur,
  Archetti, Ayhan, Battarra, Bennell, Bourjolly, Boylan
  et~al.}]{Petropoulos2024}
Fotios Petropoulos, Gilbert Laporte, Emel Aktas, Sibel~A Alumur, Claudia
  Archetti, Hayriye Ayhan, Maria Battarra, Julia~A Bennell, Jean-Marie
  Bourjolly, John~E Boylan, and 1 others. 2024.
\newblock \href {https://doi.org/10.1080/01605682.2023.2253852} {Operational
  research: methods and applications}.
\newblock \emph{Journal of the Operational Research Society}, 75(3):423--617.

\bibitem[{Qin et~al.(2024)Qin, Liang, Ye, Zhu, Yan, Lu, Lin, Cong, Tang, Qian,
  Zhao, Hong, Tian, Xie, Zhou, Gerstein, dahai li, Liu, and Sun}]{Qin2024}
Yujia Qin, Shihao Liang, Yining Ye, Kunlun Zhu, Lan Yan, Yaxi Lu, Yankai Lin,
  Xin Cong, Xiangru Tang, Bill Qian, Sihan Zhao, Lauren Hong, Runchu Tian,
  Ruobing Xie, Jie Zhou, Mark Gerstein, dahai li, Zhiyuan Liu, and Maosong Sun.
  2024.
\newblock \href {https://openreview.net/forum?id=dHng2O0Jjr} {Tool{LLM}:
  Facilitating large language models to master 16000+ real-world {API}s}.
\newblock In \emph{The Twelfth International Conference on Learning
  Representations}.

\bibitem[{Ramamonjison et~al.(2023)Ramamonjison, Yu, Li, Li, Carenini, Ghaddar,
  He, Mostajabdaveh, Banitalebi-Dehkordi, Zhou et~al.}]{ramamonjison2023nl4opt}
Rindranirina Ramamonjison, Timothy Yu, Raymond Li, Haley Li, Giuseppe Carenini,
  Bissan Ghaddar, Shiqi He, Mahdi Mostajabdaveh, Amin Banitalebi-Dehkordi,
  Zirui Zhou, and 1 others. 2023.
\newblock \href {https://proceedings.mlr.press/v220/ramamonjison23a.html}
  {Nl4opt competition: Formulating optimization problems based on their natural
  language descriptions}.
\newblock In \emph{Proceedings of the NeurIPS 2022 Competitions Track}, pages
  189--203.

\bibitem[{Romera-Paredes et~al.(2024)Romera-Paredes, Barekatain, Novikov,
  Balog, Kumar, Dupont, Ruiz, Ellenberg, Wang, Fawzi
  et~al.}]{RomeraParedes2024}
Bernardino Romera-Paredes, Mohammadamin Barekatain, Alexander Novikov, Matej
  Balog, M~Pawan Kumar, Emilien Dupont, Francisco~JR Ruiz, Jordan~S Ellenberg,
  Pengming Wang, Omar Fawzi, and 1 others. 2024.
\newblock \href {https://doi.org/10.1038/s41586-023-06924-6} {Mathematical
  discoveries from program search with large language models}.
\newblock \emph{Nature}, 625(7995):468--475.

\bibitem[{Schick et~al.(2023)Schick, Dwivedi-Yu, Dess{\`\i}, Raileanu, Lomeli,
  Hambro, Zettlemoyer, Cancedda, and Scialom}]{schick2023toolformer}
Timo Schick, Jane Dwivedi-Yu, Roberto Dess{\`\i}, Roberta Raileanu, Maria
  Lomeli, Eric Hambro, Luke Zettlemoyer, Nicola Cancedda, and Thomas Scialom.
  2023.
\newblock \href
  {http://papers.nips.cc/paper\_files/paper/2023/hash/d842425e4bf79ba039352da0f658a906-Abstract-Conference.html}
  {Toolformer: Language models can teach themselves to use tools}.
\newblock In \emph{Advances in Neural Information Processing Systems},
  volume~36, pages 68539--68551.

\bibitem[{Shi et~al.(2024)Shi, Gao, Chen, Feng, Yan, Shi, Yin, Ren, Verberne,
  and Ren}]{Shi2024}
Zhengliang Shi, Shen Gao, Xiuyi Chen, Yue Feng, Lingyong Yan, Haibo Shi, Dawei
  Yin, Pengjie Ren, Suzan Verberne, and Zhaochun Ren. 2024.
\newblock \href {https://doi.org/10.18653/v1/2024.findings-emnlp.624} {Learning
  to use tools via cooperative and interactive agents}.
\newblock In \emph{Findings of the Association for Computational Linguistics:
  EMNLP 2024}, pages 10642--10657, Miami, Florida, USA. Association for
  Computational Linguistics.

\bibitem[{Shi et~al.(2025)Shi, Gao, Yan, Feng, Chen, Chen, Yin, Verberne, and
  Ren}]{Shi2025}
Zhengliang Shi, Shen Gao, Lingyong Yan, Yue Feng, Xiuyi Chen, Zhumin Chen,
  Dawei Yin, Suzan Verberne, and Zhaochun Ren. 2025.
\newblock \href {https://doi.org/10.1145/3696410.3714825} {Tool learning in the
  wild: Empowering language models as automatic tool agents}.
\newblock In \emph{Proceedings of the ACM on Web Conference 2025}, pages
  2222--2237.

\bibitem[{Wang et~al.(2024)Wang, Fang, Eisner, Van~Durme, and Su}]{Wang2024}
Boshi Wang, Hao Fang, Jason Eisner, Benjamin Van~Durme, and Yu~Su. 2024.
\newblock \href {https://doi.org/10.18653/v1/2024.acl-long.570} {{LLM}s in the
  imaginarium: Tool learning through simulated trial and error}.
\newblock In \emph{Proceedings of the 62nd Annual Meeting of the Association
  for Computational Linguistics (Volume 1: Long Papers)}, pages 10583--10604,
  Bangkok, Thailand. Association for Computational Linguistics.

\bibitem[{Xia et~al.(2025)Xia, Li, Liu, Wu, and Liu}]{Xia2025}
Shijie Xia, Xuefeng Li, Yixin Liu, Tongshuang Wu, and Pengfei Liu. 2025.
\newblock \href {https://doi.org/10.1609/aaai.v39i26.34987} {Evaluating
  mathematical reasoning beyond accuracy}.
\newblock In \emph{Proceedings of the AAAI Conference on Artificial
  Intelligence}, pages 27723--27730.

\bibitem[{Xia et~al.(2024)Xia, Kim, Chen, Ye, Kundu, Hao, and Talati}]{Xia2024}
Yuchen Xia, Jiho Kim, Yuhan Chen, Haojie Ye, Souvik Kundu, Cong~Callie Hao, and
  Nishil Talati. 2024.
\newblock \href {https://doi.org/10.1109/IISWC63097.2024.00027} {Understanding
  the performance and estimating the cost of llm fine-tuning}.
\newblock In \emph{2024 IEEE International Symposium on Workload
  Characterization}, pages 210--223.

\bibitem[{Xiao et~al.(2023)Xiao, Zhang, Wu, Xu, Wang, Han, Fu, Zhong, Zeng,
  Song et~al.}]{Xiao2023}
Ziyang Xiao, Dongxiang Zhang, Yangjun Wu, Lilin Xu, Yuan~Jessica Wang, Xiongwei
  Han, Xiaojin Fu, Tao Zhong, Jia Zeng, Mingli Song, and 1 others. 2023.
\newblock \href {https://openreview.net/forum?id=HobyL1B9CZ} {Chain-of-experts:
  When llms meet complex operations research problems}.
\newblock In \emph{The twelfth international conference on learning
  representations}.

\bibitem[{Yang et~al.(2023)Yang, Song, Li, Zhao, Ge, Li, and Shan}]{Yang2023}
Rui Yang, Lin Song, Yanwei Li, Sijie Zhao, Yixiao Ge, Xiu Li, and Ying Shan.
  2023.
\newblock \href
  {http://papers.nips.cc/paper\_files/paper/2023/hash/e393677793767624f2821cec8bdd02f1-Abstract-Conference.html}
  {Gpt4tools: Teaching large language model to use tools via self-instruction}.
\newblock In \emph{Advances in Neural Information Processing Systems}, pages
  71995--72007.

\bibitem[{Ye et~al.(2024{\natexlab{a}})Ye, Li, Li, Huang, Gao, Wu, Zhang, Gui,
  and Huang}]{Ye2024a}
Junjie Ye, Sixian Li, Guanyu Li, Caishuang Huang, Songyang Gao, Yilong Wu,
  Qi~Zhang, Tao Gui, and Xuanjing Huang. 2024{\natexlab{a}}.
\newblock \href {https://doi.org/10.18653/v1/2024.acl-long.119} {{T}ool{S}word:
  Unveiling safety issues of large language models in tool learning across
  three stages}.
\newblock In \emph{Proceedings of the 62nd Annual Meeting of the Association
  for Computational Linguistics (Volume 1: Long Papers)}, pages 2181--2211,
  Bangkok, Thailand. Association for Computational Linguistics.

\bibitem[{Ye et~al.(2024{\natexlab{b}})Ye, Wu, Gao, Huang, Li, Li, Fan, Zhang,
  Gui, and Huang}]{Ye2024}
Junjie Ye, Yilong Wu, Songyang Gao, Caishuang Huang, Sixian Li, Guanyu Li,
  Xiaoran Fan, Qi~Zhang, Tao Gui, and Xuanjing Huang. 2024{\natexlab{b}}.
\newblock \href {https://doi.org/10.18653/v1/2024.emnlp-main.19} {{R}o{TB}ench:
  A multi-level benchmark for evaluating the robustness of large language
  models in tool learning}.
\newblock In \emph{Proceedings of the 2024 Conference on Empirical Methods in
  Natural Language Processing}, pages 313--333, Miami, Florida, USA.
  Association for Computational Linguistics.

\bibitem[{Zhang et~al.(2024)Zhang, Liu, Cherry, and Firat}]{Zhang2024}
Biao Zhang, Zhongtao Liu, Colin Cherry, and Orhan Firat. 2024.
\newblock \href {https://openreview.net/forum?id=5HCnKDeTws} {When scaling
  meets {LLM} finetuning: The effect of data, model and finetuning method}.
\newblock In \emph{The Twelfth International Conference on Learning
  Representations}.

\bibitem[{Zheng et~al.(2024)Zheng, Li, Yan, Zhang, Huang, and Liu}]{Zheng2024}
Yuanhang Zheng, Peng Li, Ming Yan, Ji~Zhang, Fei Huang, and Yang Liu. 2024.
\newblock \href {https://doi.org/10.18653/v1/2024.findings-acl.536}
  {Budget-constrained tool learning with planning}.
\newblock In \emph{Findings of the Association for Computational Linguistics:
  ACL 2024}, pages 9039--9052, Bangkok, Thailand. Association for Computational
  Linguistics.

\end{thebibliography}

\clearpage
%
%
%
%
%
%
%
%

\appendix
\section{Question and Answer Generation Prompts}

\subsection{Problem Generation Prompt Template}\label{sec:question-prompt}
\begin{figure}[h]
	\centering
	\includegraphics[width=\linewidth]{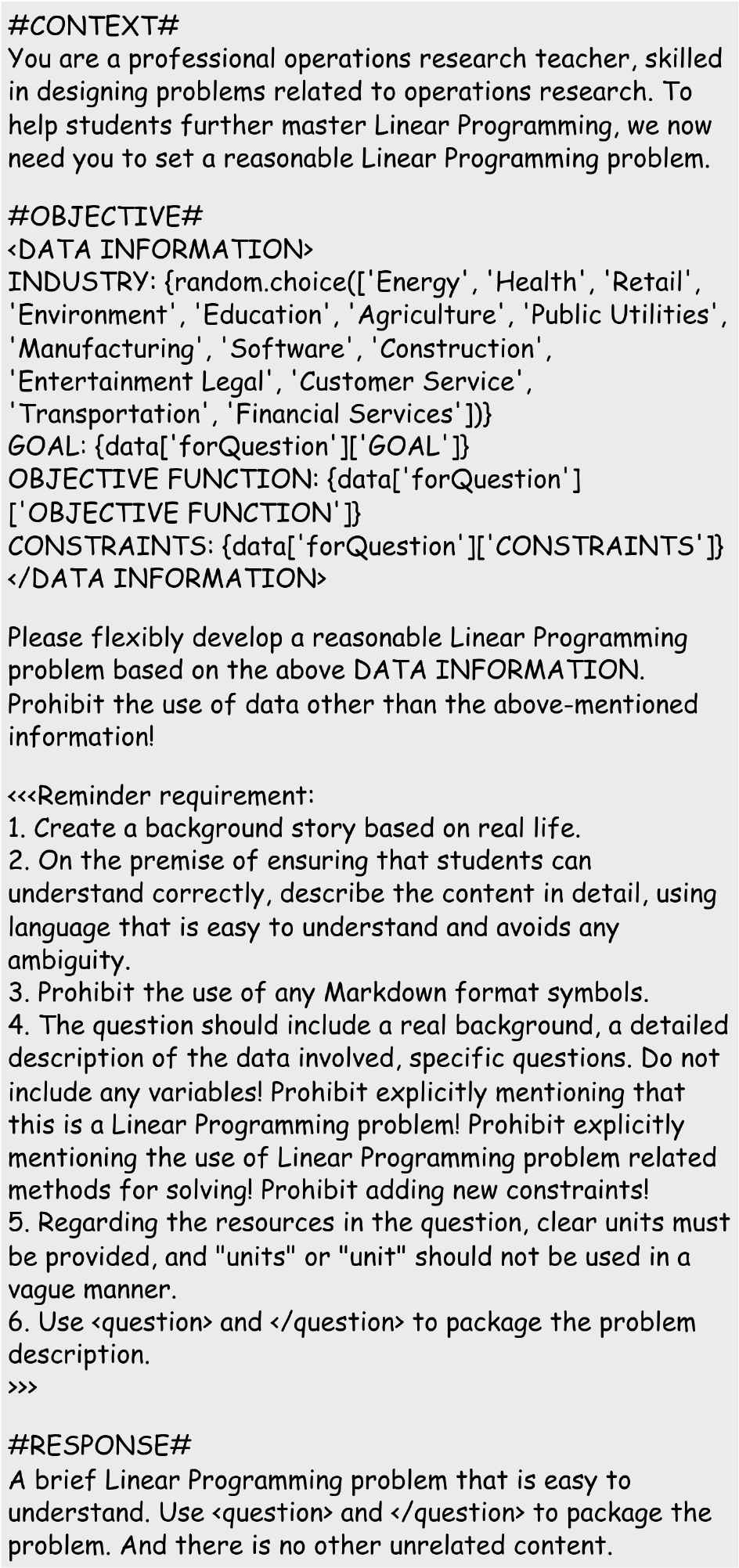}
	\caption{Problem generation prompt template.}
	\label{fig:question-prompt}
\end{figure}

Figure~\ref{fig:question-prompt} shows the prompt template for generating linear programming problem statements, with key information of problems (as illustrated in Figure~\ref{fig:data-generation}) inserted into \{\}.

\subsection{Answer Generation Prompt Template}\label{sec:answer-prompt}
\begin{figure}[!h]
	\centering
	\includegraphics[width=\linewidth]{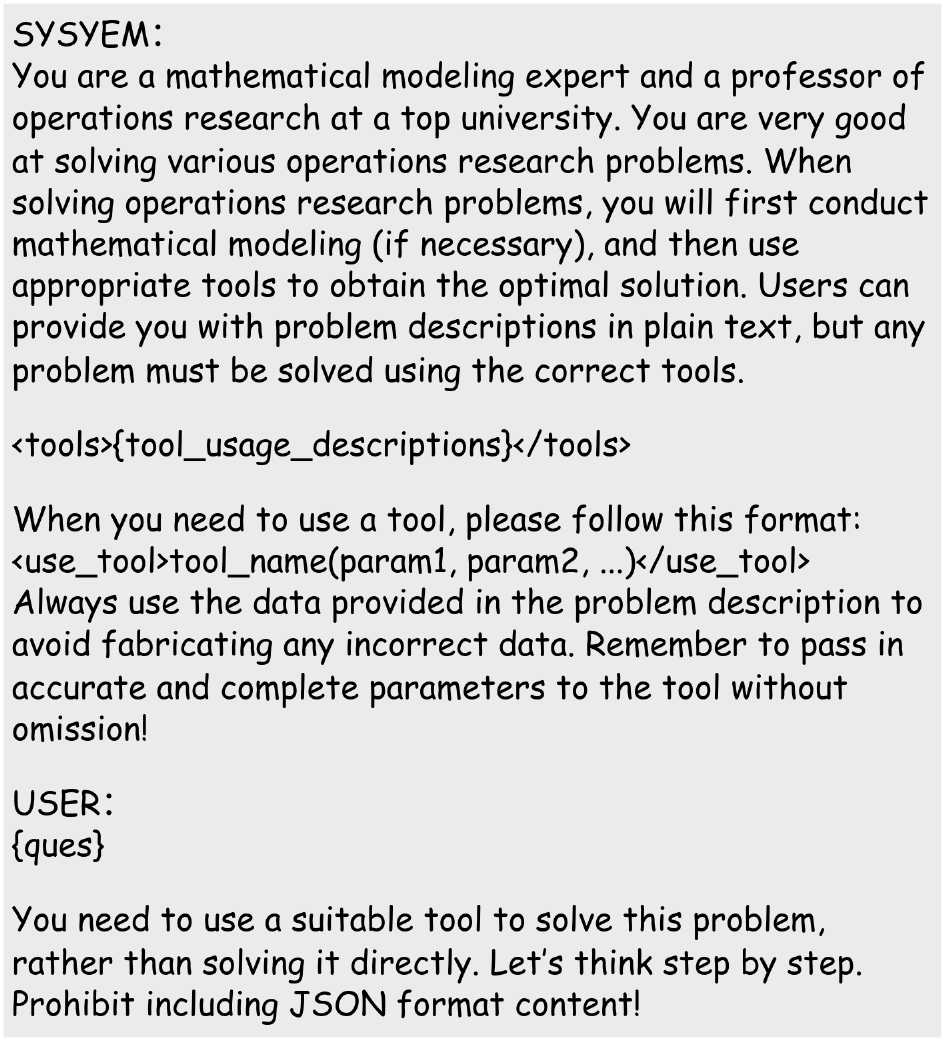}
	\caption{Answer generation prompt template.}
	\label{fig:answer-prompt}
\end{figure}

Figure~\ref{fig:answer-prompt} shows the prompt template for generating answers, with the API usage description and OR question statements (as illustrated in Figure~\ref{fig:data-generation}) inserted into \{\}. This template is used to generate input for OR-Toolformer and all baselines.

\section{Data, Code, and Model Availability}

The dataset, source code, and model checkpoints are available at the anonymized URL for peer review:
\url{https://figshare.com/s/262251a08ea7f79113d7}.

\end{document}